\documentclass{llncs}
\usepackage[latin1]{inputenc}
\usepackage{times}
\usepackage{helvet}
\usepackage{courier}
\usepackage{calc,graphicx,latexsym,relsize,url,xspace}
\usepackage{algorithm}                 % to get the algorithm in a figure
\usepackage{xypic}
\usepackage{subfigure}

\newcommand{\comment}[1]{}

\input{epsf}

\def\0{{\bf 0}}
\def\1{{\bf 1}}

\newcommand{\alg}[1]{\textsf{\smaller{#1}}}

\newcommand\Omit[1]{}

\begin{document}

\title{Elicitation strategies for fuzzy constraint problems with missing 
preferences: algorithms and experimental studies}
%Reasoning about fuzzy constraint problems with
%missing preferences: an experimental study}
	%:\\ a general framework}
\author{
Mirco Gelain\inst{1}, Maria Silvia Pini\inst{1},  
Francesca Rossi\inst{1}, K. Brent Venable\inst{1}, and Toby Walsh\inst{2}} 
\institute{
Dipartimento di Matematica Pura ed Applicata,
Universit\`a di Padova, Italy\\
E-mail: \{mgelain,mpini,frossi,kvenable\}@math.unipd.it 
\and 
 NICTA and UNSW Sydney, Australia, \\ Email: Toby.Walsh@nicta.com.au
}

%\author{Confirmation number: 91
%\institute{ }}
\date{ }

\maketitle

\bibliographystyle{plain}

\begin{abstract}
Fuzzy constraints are a popular approach to handle preferences and
over-constrained problems
in scenarios where one needs to be cautious, such as in medical or space
applications.
We consider here fuzzy constraint problems where some of the preferences
may be missing.
This models, for example, settings where agents are distributed and have
privacy issues, or where there is an ongoing preference elicitation
process.
In this setting, we study how to find a solution which is optimal
irrespective
of the missing preferences. In the process of finding such a solution,
we may elicit preferences from the user if necessary. However, our goal
is
to ask the user as little as possible.
We define a combined solving and preference
elicitation scheme with a large number of different instantiations, each
corresponding to a concrete algorithm which we compare experimentally. We
compute both the number of elicited preferences and the "user effort",
which may be larger, as it contains all the preference values the user
has to compute to be able to respond to the elicitation requests. While
the number of elicited preferences is important when the concern
is to communicate as little information as possible, the user effort
measures also the hidden work the user has to do to be able to
communicate the elicited preferences. 
Our experimental results show that some of our algorithms are very good
at
finding a necessarily optimal solution while asking the user for only a
very small
fraction of the missing preferences.
The user effort is also very small for the best algorithms.
Finally, we test these algorithms on hard constraint problems
with possibly missing constraints, where the aim is to find feasible
solutions irrespective of the missing constraints.
\end{abstract}

\section{Introduction}

Constraint programming is a powerful
paradigm for solving scheduling, planning, and resource allocation 
problems. A problem is represented by a set of variables, 
each with a domain of values, and a set of constraints. A solution 
is an assignment of values to the variables 
which satisfies all constraints and
which optionally maximizes/minimizes an objective function.
Soft constraints are a way to model optimization problems by allowing for 
several levels of satisfiability, modelled by the use of preference or cost values
that represent how much we like an instantiation of
the variables of a constraint.

It is usually assumed that the data
(variables, domains, (soft) constraints) is completely known before 
solving starts. This is often unrealistic. 
In web applications and multi-agent systems, 
the data is frequently only partially known 
and may be added to at a later date by, for example, elicitation.
Data may also come from different sources
at different times. 
In multi-agent systems, agents may release data
reluctantly due to privacy concerns.

%Several approaches have tried to address these issues:
%For example, open CSPs \cite{open1,open2} and 
%interactive CSPs \cite{interactive}
%work with domains that can be partially specified.
%As a second example, in dynamic CSPs \cite{dynamic} variables, domains, 
%and constraints may change over time.

%It has been shown that these approaches are closely related.
%In fact, interactive CSPs can be seen as a special case of 
%both dynamic and open CSPs \cite{pedro}.
%thus solvers developed for dynamic 
%and open CSPs can be used also for interactive CSPs
%\cite{pedro}.
%

Incomplete soft constraint problems can model
such situations by allowing some of the preferences 
to be missing. An algorithm has been proposed and tested to solve 
such incomplete problems \cite{cp07-isoft}. 
The goal is to find a solution 
that is guaranteed to be optimal irrespective of
the missing preferences, eliciting
preferences if necessary until 
such a solution exists. 
Two notions of optimal solution are considered:
{\em possibly optimal} solutions are assignments 
that are optimal in {\em at least one way} of
revealing the unspecified preferences, while 
{\em necessarily optimal} solutions are assignments 
that are optimal in {\em all ways} that the
unspecified preferences can be revealed. 
The set of possibly optimal solutions is never empty, while 
the set of necessarily optimal solutions can be empty. 

If there is no necessarily optimal solution,
the algorithm proposed in \cite{cp07-isoft} uses branch and bound 
to find a "promising solution" (specifically, a complete assignment 
in the best possible completion of the current problem) 
and elicits the missing preferences
related to this assignment. This process is repeated till
there is a necessarily optimal solution. 

Although this algorithm behaves reasonably well,
it make some specific choices about solving and 
preference elicitation that may not be optimal in practice,
as we shall see in this paper. For example, the algorithm
only elicits missing preferences after running branch
and bound to exhaustion. As a second example, the 
algorithm elicits all missing preferences related to
the candidate solution. 
Many other strategies are possible.
We might elicit preferences at the end of every complete branch,
or even at every node in the search tree. 
Also, when choosing the value to assign to a variable, we might
ask the user (who knows the missing preferences) for help.
Finally, we might not elicit all the missing preferences 
related to the current candidate solution.
For example, we might just ask the user for
the worst preference among the missing ones. 

In this paper we consider a general algorithm scheme 
which greatly generalizes that proposed in \cite{cp07-isoft}.
It is based on three parameters:
{\em what} to elicit, {\em when} to elicit it, and {\em who} chooses 
the value to be assigned to the next variable. 
We test all 16 possible different instances of the scheme 
(among which is the algorithm in \cite{cp07-isoft})
on randomly generated fuzzy constraint problems.
We demonstrate that some of the algorithms are very good at 
finding necessarily optimal solution without eliciting too many 
preferences. %For example, .....
We also test the algorithms on problems with hard constraints. 
%obtaining even better results.
Finally, we consider problems with fuzzy temporal constraints,
where problems have more specific structure.

In our experiments, we compute the %percentage of 
elicited preferences, that is, the missing values that the user has to provide to the system 
because they are requested by the algorithm.
Providing these values usually has a cost, either in terms of computation effort, or in 
terms of privacy decrease, or also in terms of communication bandwidth. 
Thus knowing how many preferences are elicited is important if we care about any of these 
issues.
However, we also compute a measure of 
the user's effort, which may be larger than the number of elicited preferences, 
as it contains all the preference values the user
has to consider to be able to respond to the elicitation requests. 
For example, we may ask the user for the worst preference value among $k$ missing ones:
the user will communicate only one value, but he will have to consider all $k$ of them.
While knowing the number of elicited preferences is important when the concern
is to communicate as little information as possible, the user effort
measures also the hidden work the user has to do to be able to
communicate the elicited preferences.
This user's effort is therefore also an important measure.

As a motivating example, recommender systems give suggestions based on
partial knowledge of the user's preferences. Our approach could
improve performance by identifying some key questions to ask before
giving recommendations. Privacy concerns regarding the percentage of
elicited preferences are motivated by eavesdropping. User's effort is
instead related to the burden on the user.

Our results show that the choice of preference elicitation strategy is 
crucial for the performance of the solver.
While the best algorithms need to elicit as little as 10\% of 
the missing preferences, the worst one needs much more.
The user's effort is also very small for the best algorithms.
The performance of the best algorithms shows that 
we only need to ask the user a very small amount of additional 
information to be able to solve problems with missing data.

Several other approaches have addressed similar issues.
For example, open CSPs \cite{open1,open2} and 
interactive CSPs \cite{interactive}
work with domains that can be partially specified.
As a second example, in dynamic CSPs \cite{dynamic} variables, domains, 
and constraints may change over time.
However, the incompleteness considered
in \cite{open2,open3} is on domain values as well as on their preferences. % or costs.
Working under this assumption means
that the agent that provides new values/costs
for a variable knows all possible costs, since they are capable of providing
the best value first.
If the cost computation is expensive or time consuming,
then computing all such costs (in order to give the
most preferred value) is not desirable.
%This is not needed in our setting, where single preferences are elicited.
We assume instead, as in \cite{cp07-isoft},
that all values are given at the beginning, and that
only some preferences are missing. 
Because of this assumption,
we don't need to elicit preference values in order,
as in \cite{open2}.

\section{Background}

In this section we give a brief overview of the fundamental notions and 
concepts on Soft Constraints and Incomplete Soft Constraints.  

Incomplete Soft Constraints problems (ISCSPs) \cite{cp07-isoft} extend 
Soft Constraint Problems (SCSPs) \cite{jacm} to deal 
with partial information. We will focus on  
a specific instance of this framework
%Incomplete {\em Fuzzy} Constraint Problems (IFCSPs)
in which
the soft constraints are fuzzy.

%and thus in what follows ISCSP 
%will indicate such a kind of problem.  
%
%More precisely
%\begin{definition}[incomplete soft constraint]
Given a set of  variables $V$ with finite domain $D$,
%and a c-semiring $\langle A, +, \times, 0,1\rangle$,
an {\em incomplete fuzzy constraint} %, or simply an {\em incomplete soft constraint},
is a pair $\langle idef,$ $con\rangle$ 
where $con \subseteq V$ is the scope of the constraint and 
$idef: D^{|con|} \longrightarrow [0,1] \cup \{?\}$ is the 
preference function of the constraint associating to each tuple 
of assignments to the variables in $con$ either 
a preference value ranging between 0 and 1, or $?$.
All tuples mapped into $?$ by $idef$ are called {\em incomplete tuples}, 
meaning that their preference is unspecified.
%\end{definition}
%In an incomplete soft constraint, the preference function can  
%either specify the preference value of a tuple by assigning a specific element
%from the carrier of the c-semiring, or leave such preference unspecified. 
%Formally, in the latter case the associated value is $?$.
A fuzzy constraint %\cite{fuzzy} 
is an incomplete fuzzy constraint with no incomplete tuples.
%where all the tuples have a specified preference. 

%\begin{definition}[incomplete soft constraint problem (ISCSP)]    
An {\em incomplete fuzzy constraint problem} (IFCSP) is a pair     
$\langle C, V, D \rangle$ where $C$ is a set of incomplete fuzzy constraints    
over the variables in $V$ with domain $D$.   
Given an IFCSP $P$, $IT(P)$ denotes the set of all incomplete tuples in $P$.   
When there are no incomplete tuples, we will denote a 
fuzzy constraint problem by FSCP.
%rop the I from the notation 
%\end{definition}      
   
%Moreover,
%\begin{definition}[completion]   
Given an IFCSP $P$, a {\em completion of $P$} is an IFCSP $P'$ 
obtained from $P$    
by associating to each incomplete tuple in every constraint    
an element in $[0,1]$.   
A completion is {\em partial} if some preference remains unspecified.   
%completion is an ISCSP $P'$ obtained from $P$    
%by associating to some incomplete tuple in some constraints    
%an element of the carrier of the c-semiring.   
$C(P)$ denotes the set of all possible completions of $P$   
and $PC(P)$ denotes the set of all its partial completions.    
Given an assignment $s$ to all the variables of an IFCSP $P$, 
$pref(P,s)$ is the preference of $s$ in $P$, defined as %. More precisely,
$pref(P,s)= min_{<idef,con> \in C | idef(s_{\downarrow con}) \neq ?} 
idef(s_{\downarrow con})$. 
%where $s_{\downarrow con}$ is the projection of 
%$s$ on the variables in $con$.
It is obtained 
%The preference of an assignment $s$ in an incomplete 
%fuzzy constraint problem $P$ is obtained 
by taking the minimum among the 
known preferences associated to the projections 
of the assignment, that is, of the appropriated sub-tuples in the constraints. 
%The projections which have unspecified preferences
%are simply ignored. 
%It is easy to see that $pref(P,s)$ is an upper bound to the preference
%of $s$ in any completion of $P$. In fact, by adding some preferences to the problem, 
%the preference of an assignment can only decrease or remain the same.

%\begin{example}
%\label{ex3}
%Consider the two assignments $s_1 = (p,m,b)$
%and $s_2 = (p,m,su)$,
%we have that $pref(P,s_1)=min(0.8,0.7,0.9,0.2)=0.2$, while
%$pref(P,s_2)= min($ $0.8,0.7,0.9)=0.7$.
%However, while the preference of $s_1$ is fixed, 
%since none of its projections is incomplete, the preference of 
%$s_2$ may become lower that $0.7$ depending on 
%the preference of the incomplete tuple $(su,m)$.    
%\end{example}

%As shown by the example, the presence of incompleteness generates a 
%partition of the set of assignments into two sets: those which have 
%a certain preference which is independent of how incompleteness 
%is resolved, and those whose preference 
%is only an upper-bound, in the sense that it can be lowered in some completions.

%Given an ISCSP $P$, we will denote the first set of assignments 
%as $Fixed(P)$ and the second with $Unfixed(P)$.
%In Example \ref{ex3}, $Fixed(P)=\{s_1\}$, while
%all other assignments belong to $Unfixed(P)$.

In the fuzzy context, a complete assignment of values to all the variables
is an optimal solution if its %global 
preference is maximal. % among those of other solutions. 
The optimality notion of FCSPs is 
generalized to IFCSPs via the notions of 
{\em necessarily and possibly optimal solutions}, that is, 
complete assignments which are maximal in all or some completions.
%More precisely, 
%given an IFCSP $P=\langle C, V, $ $D \rangle$, 
%an assignment $s \in D^{|V|}$ 
%is a necessarily (resp, possibly) optimal solution for $P$
%iff $\forall Q \in C(P)$ (resp., $\exists Q \in C(P)$ such that) 
%$\mbox{ }\forall s'\in D^{|V|}$,  $pref(Q,s') \not > pref(Q,s)$.
Given an IFCSP $P$, we denote by $NOS(P)$ (resp., $POS(P)$)
the set of necessarily (resp., possibly) optimal solutions of $P$. 
Notice that $NOS(P) \subseteq POS(P)$. Moreover, 
while $POS(P)$ is never empty, $NOS(P)$ may be empty. 
In particular, $NOS(P)$ is empty whenever the revealed preferences 
do not fix the relationship between one assignment and all
others. 

%Solving an IFCSP, thus, means finding an necessarily optimal solution 
%while asking the user to reveal the smallest number of incomplete tuples.   
%Such a solution, in fact is robust w.r.t. the missing preferences since 
%it is optimal in all the completions.

%\begin{example}
%In the ISCSP $P$ of Figure \ref{agency3}, 
%we can easily see that $NOS(P)= \emptyset$ 
%since, given any assignment, it is 
%possible to construct a completion of $P$ in which 
%it is not an optimal solution. 
%On the other hand, $POS(P)$ contains all assignments not 
%including tuple $(sh,m)$. 
%%In fact, such a tuple
%%has preference $0.1$ and it 
%%drowns the preference of any 
%%assignment containing it below the preference of fixed solution 
%%$s_1=(p,m,b)$ (i.e., below $0.2$). Thus, in all completions, $s_1$ dominates 
%%any such assignment.
%%Instead, for any assignment $s$ not including $(sh,m)$, 
%%we can complete $P$ in order to make $s$ optimal, for example by setting 
%%the preferences of all the incomplete tuples of $s$ to $pref(s,P)$ and 
%%the preferences of all other incomplete tuples to 0.   
%\end{example}

%In \cite{isoft-cp07} a characterization of  
%\section{Characterizing POS(P) and NOS(P)}
%\label{charact}

In \cite{cp07-isoft} an algorithm is proposed 
to find a necessarily optimal solution of an IFCSP 
based on a characterization of $NOS(P)$ and $POS(P)$.
This characterization uses the preferences of the optimal solutions
%namely $pref_0$ and $pref_1$, 
of two special completions of $P$, namely 
the  $\0$-completion of $P$, denoted by $P_0$, % \in C(P)$, 
obtained from $P$ by associating preference $0$ to each tuple of $IT(P)$, and the 
$\1$-completion of $P$, denoted by $P_1$, % \in C(P)$, 
obtained from $P$ by associating preference $1$ to each tuple of $IT(P)$.
Notice that, by monotonicity of $min$, we have that $pref_0 \leq pref_1$.
When $pref_0 = pref_1$, $NOS(P) = Opt(P_0)$; thus, any optimal solution 
of $P_0$ is a necessary optimal solution. 
Otherwise, $NOS(P)$ is empty and $POS(P)$ is a set of solutions
with preference between $pref_0$ and $pref_1$ in $P_1$.
The algorithm proposed in \cite{cp07-isoft} 
finds a necessarily optimal solution of the given IFCSP by 
interleaving the computation of $pref_0$ and $pref_1$ 
with preference elicitation steps, until the two values coincide. Moreover, 
the preference elicitation is guided by the fact that only solutions in $POS(P)$ 
%those having preference in  $P_1$ that lies between $pref_0$ and $pref_1$, 
can become necessarily optimal. Thus, the algorithm only elicits 
preferences related to optimal solutions of $P_1$. 
 
%\section{A general solver scheme and its instances}
\section{A general solver scheme}
%New Algorithms for finding robust solutions}
\label{solving}

We now propose a more general schema for solving IFCSPs 
based on interleaving branch and bound (BB) search with elicitation. 
This schema generalizes the concrete solver presented in \cite{cp07-isoft},
but has several other instantiations that 
we will consider and compare experimentally in this paper.
The scheme uses branch and bound. 
This considers the variables in some order, 
choosing a value for each variable, and pruning
branches based on an upper bound (assuming the goal is to maximize)
on the preference value 
of any completion of the current partial assignment. 
To deal with missing preferences, 
branch and bound
is applied to both the 0-completion and 
the $\1$-completion of the problem.  
If they have the same solution,
this is a necessarily optimal solution and we can stop. If not, 
we elicit some of the missing preferences and
continue branch and bound on the new $\1$-completion. 

Preferences can be elicited after each run of branch and bound
(as in \cite{cp07-isoft}) or during a BB run while preserving the 
correctness of the approach.
For example,
we can elicit preferences at the end of every complete branch
(that is, regarding preferences of every complete assignment considered in the
branch and bound algorithm), or at every node in the search tree
(thus considering every partial assignment).
Moreover, when choosing the value for the next variable to be assigned,
we can ask the user (who knows the missing preferences) for help.
Finally, rather than eliciting all the missing preferences in the
possibly optimal solution, or the complete or partial
assignment under consideration, 
we can elicit just one of the missing preferences. For example, 
with fuzzy constraint problems,
eliciting just the worst preference among the missing 
ones is sufficient since only the worst value is important
to the computation of the overall preference value.
More precisely, the algorithm schema we propose is
based on the following parameters:

\begin{enumerate}
\item \textbf{Who} chooses the value of a variable:   
%\begin{enumerate}
%\item 
the algorithm can choose the values in decreasing order either 
w.r.t. their preference values in the $\1$-completion (Who=dp)
or in the $\0$-completion (Who=dpi).
Otherwise, the user can suggest this choice. To do this, he can consider 
all the preferences (revealed or not) for the values 
of the current variable
%that is the most preferred among those in the domain 
%which haven't been assigned yet 
({\em lazy user}, Who=lu for short);
or he considers also the preference
values in constraints between this 
variable and the past variables in the search order
%the value that is the most preferred among 
%those which haven't been assigned yet given the constraints 
%involving the variable being assigned and the (past) instantiated variables 
({\em smart user}, Who=su for short). 
%%one of  ``medium level'', that considers also 
%\end{enumerate}
%\end{enumerate}

\item \textbf{What} is elicited: 
%given a possibly partial assignment
%it is possible to ask the user to reveal 
%\begin{enumerate}
%\item 
we can elicit 
the preferences of all the incomplete tuples of the current assignment 
(What=all) 
or only the worst preference in the current assignment, 
if it is worse than the known ones (What=worst);
%\item only the preference of the best tuple of the current assignment 
%which is lower than the current lower bound, 
%if it exists (denoted with {\em HIGHEST\_LESS\_THAN\_LB}).
%\end{enumerate}

\item \textbf{When} elicitation takes place: 
we can elicit preferences at the end of the branch
and bound search (When=tree), 
or during the search, 
when we have a complete assignment to all variables
(When=branch) 
or whenever a new value is assigned to a variable (When=node).
\end{enumerate}

\begin{figure}[h]
\begin{center} 
%{	\centering
		%\includegraphics[width=0.4\textwidth]{./grafici/grafici/diagramma1_alg}
			\includegraphics[width=0.6\textwidth]{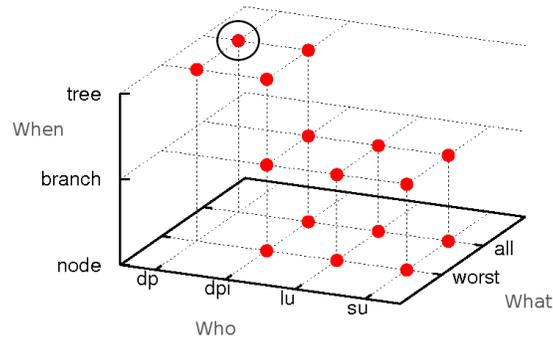}
%}
\end{center}
	\caption{Instances of the general scheme.}
%The solvers considered in this paper, as instances of the general scheme.} 
%algorithms. Circled is the one in \cite{cp07-isoft}.}
	\label{fig:var-alg}
\end{figure}

By choosing a value for each of the three above 
parameters in a consistent way, 
we obtain in total 16 different algorithms, as summarized 
%in the figure below, where 
in Figure \ref{fig:var-alg}, where 
the circled instance is the concrete solver used in \cite{cp07-isoft}.

%Every point in Figure \label{fig:var-alg} represents an
%instantiation of Find-NOS to specific values 
%for parameters  $who$, $what$ and $when$.

\begin{figure}[htbp]
  \begin{tabular}{l}\hline
  \begin{minipage}{2.2in}
  \begin{footnotesize}
  \begin{tabbing}
  \alg{IFCSP-SCHEME}($P$,$Who$,$What$,$When$)\\  
  %{\bf FIND-NOS}($P$, $Who$,$What$,$When$)\\   
       %$P_0 \leftarrow P[?/0]$, 
$Q \leftarrow P_0$\\
$s_{max}$, $pref_{max}$ $\leftarrow BB(P_0,-)$\\
 $Q'$,$s_1$,$pref_1$ $\leftarrow BBE(P, 0, Who, What, When, s_{max},pref_{max})$\\
  If \= ($s_1 \neq nil$)\\
 	\>	$s_{max} \leftarrow s_1$, $pref_{max} \leftarrow pref_1$, $Q \leftarrow Q'$\\
Return $Q$, $s_{max}$, $pref_{max}$
 \end{tabbing}
  \end{footnotesize}
  \end{minipage}
  \\\hline
  \end{tabular}
\caption{Algorithm IFCSP-SCHEME.} 
\label{alg1}
\end{figure}

Figures \ref{alg1} and \ref{alg2} show the 
pseudo-code of the general scheme for solving IFCSPs. There
are three algorithms: ISCSP-SCHEME, BBE and BB. 
%is given in Figure Algorithms Find-NOS and BBE. 
%la pref ritornata dalle procedure elicit e' la preferenza dell'assegnamento 
%globale o parziale corrente solo se fuzzy quando what=worst
ISCSP-SCHEME takes 
as input an IFCSP $P$ and the values 
for the three parameters: Who, What and When.
It returns a partial completion of $P$ that has some necessarily optimal solutions,
one of these necessarily optimal solutions, and its preference value. 
It starts by %assigning $P_0$ to the 0-completion of $P$, and by 
computing via branch and bound (algorithm BB)
an optimal solution of $P_0$, say $s_{max}$,
and its preference $pref_{max}$.
Next, procedure $BBE$ is called. If $BBE$ succeeds, it returns 
a partial completion of $P$, say $Q$, 
one of its necessarily optimal solutions, say $s_1$,
and its associated preference $pref_1$.
Otherwise, it returns a solution equal to $nil$. 
In the first case the output of 
IFCSP-SCHEME coincides with that of BBE, otherwise IFCSP-SCHEME
returns $P_0$, one of its optimal solutions, and its preference.

Procedure BBE takes as input the same values as IFCSP-SCHEME and, 
in addition, a solution $sol$ 
and a preference $lb$ representing the 
current lower bound on the optimal preference value. % $lb$.
Function $nextVariable$, applied to the $\1$-completion of the IFCSP, 
returns the next variable to be assigned.  
The algorithm then assigns a value to this variable. If 
the Boolean function $nextValue$ returns true (if there is a value in the domain),
we select a value for $currentVar$ according to the 
value of parameter $Who$.

Function $UpperBound$ computes an upper bound 
on the preference of any completion of the current 
partial assignment:  %In general, many kinds of upper bound can be used.
%We have chosen 
the minimum over the preferences of the constraints involving only 
variables that have already been instantiated. 
%\footnote{Per altri possibili metodi di calcolo dell'upper bound si veda la sezione \ref{subsec:Solver}}. %Formally, let $t$ be the current partial assignment to 
%variables in $\{v_1, \dots, v_k\} \subseteq V$, 
%and let  $c_i = \langle def_i,con_i\rangle$ 
%be a constraint, such that $con_i \subseteq \{v_1, \dots, v_k\}$.
%Then, the value $ub$ returned by $UpperBound$
%is:  
%\begin{equation*}
%$ub =  min_{i=1}^k def_i(t \downarrow_{con_i})$. 
%\end{equation*}
%where $\prod$ is the combination operator of the semiring.
  
%We will now describe procedure BBE  
%by considering the value of parameter $when$ fixed. 
%This corresponds to considering the algorithms 
%in Figure \ref{fig:var-alg} 
%divided into the three planes obtained 
%fixing the value on the $when$-axis.   

If When=tree, 
elicitation is handled by procedure $Elicit@tree$,
and takes place only at the end of the search over the $\1$-completion. 
The user is not involved in the value assignment steps within the search.
At the end of the search, if a solution is found, the user 
is asked either to reveal all the preferences 
of the incomplete tuples in the solution (if What=all), 
or only the worst one among them (if What=worst). 
If such a preference is better than the best found so far, BBE is called 
recursively with the new best solution and preference.

If When=branch, BB is performed only once.
The user may be asked to 
choose the next value for the current variable being instantiated.
Preference elicitation, which is handled by function $Elicit@branch$,
takes place during search, whenever all variables have been instantiated and 
the user can be asked either to reveal 
the preferences of all the incomplete tuples in the assignment (What=all),
or the worst preference among those of the 
incomplete tuples of the assignment
%, but only if such a preference 
%is lower than the preferences of the known tuples 
(What=worst).  
In both cases the information gathered is sufficient to 
test such a preference value against the current lower bound.

If When=node,
preferences are elicited every time a new value is assigned to 
a variable and it is handled by procedure $Elicit@node$. 
The tuples to be considered for elicitation 
are those involving the value which has just been assigned 
and belonging to constraints between the current variable and already 
instantiated variables. 
If What=all, the user is asked to provide the preferences of 
all the incomplete tuples involving the new assignment.
Otherwise if What=worst, the user provides only 
the preference of the worst tuple.
%, if it is lower than the known preferences. 

%It can be shown that 
%Algorithm IFCSP-SCHEME always 
%terminates and is correct.
% w.r.t. finding a necessarily optimal solution of 
%a partial completion of the IFCSP given in input.

\begin{figure}[htbp]
  \begin{tabular}{l}\\\hline
  \begin{minipage}{2.2in}
  \begin{footnotesize}
  \begin{tabbing}
\alg{BBE} \= ($P$,$nInstVar$, $Who$, $What$, $When$, $sol$, $lb$)\\
 $sol'$ $\leftarrow$ $sol$, $pref'$  $\leftarrow$ $lb$\\
 $currentVar \leftarrow nextVariable(P_1)$\\
 While \= ($nextValue(currentVar, Who)$)\\
 \> If \= ($When = node$)\\
 \> \>      $P,pref \leftarrow Elicit@Node(What, P, currentVar, lb)$\\
 \>     $ub \leftarrow UpperBound(P_1, currentVar)$\\
 \>     If \= ($ub\ >\ lb$)\\
 \> \>          If \= ($nInstvar$ = $number\ of\ variables\ in\ P$)\\
 \> \> \>                 If \= ($When$ = $branch$)\\
 \> \> \> \>                             $P,\ pref \leftarrow Elicit@branch(What, P, lb)$\\
 \> \> \>                 If \= ($pref\ >\ lb$)\\
 \> \> \> \>                             $sol \leftarrow getSolution(P_1)$\\
 \> \> \> \>                             $lb \leftarrow pref(P_1, sol)$\\
 \> \>          else \\   %else
 \> \> \>                     $BBE(P,nInstVar+1,Who,What,When,sol,lb)$\\
 If \= ($When$=$tree$ $and$ $nInstVar=0$)\\
 \>     If\= ($sol$ = $nil$)\\
  \> \>           $sol$ $\leftarrow$ $sol'$, $pref$ $\leftarrow$ $pref'$\\
 \>     else \\
 \> \>         $P,\ pref \leftarrow Elicit@tree(What, P, sol, lb)$\\
 \> \>         If\= ($pref > pref'$) \\
 \> \> \>                    $BBE(P,0,Who,What,When,sol,pref)$\\
 \> \>        else  $BBE(P,0,Who,What,When,sol',pref')$
  \end{tabbing}
  \end{footnotesize}
  \end{minipage}
  \\\hline
  \end{tabular}
\caption{Algorithm BBE.}
\label{alg2}
\end{figure}

\begin{theorem}
Given an IFCSP $P$ and a consistent set of values for parameters When, 
What and Who, Algorithm {\em IFCSP-SCHEME} always terminates, and returns 
an IFCSP $Q \in PC(P)$, an assignment $s \in NOS(Q)$, 
and its preference in $Q$.
\end{theorem}

%The proof is omitted due to lack of space.

%-----------------------------
%\begin{theorem}
\proof{Let us first notice that, as far as correctness and termination concern, the value of parameter Who is irrelevant.

We consider two separate cases, i.e., When=tree and
and When=branch or node.  

%Given an ISCSP $P$ and $when = tree$ in input, 
%then if 
%\begin{itemize}
%\item $what = all$ or
%\item $what = worst$ and $P$ a is a fuzzy ISCSP
%\end{itemize} 
%algorithm {\em Find-NOS} always terminates and returns
%an ISCSP $Q$ such that $Q \in PC(P)$, 
%an assignment $s \in NOS(Q)$, and its preference 
%in $Q$.
%\end{theorem} 

%\proof{
%Termination.

{\em Case 1:} When =tree.\\
Clearly IFCSP-SCHEME terminates if and only if  BBE terminates.
If we consider the pseudocode of procedure BBE shown in Algorithm 
\ref{alg2}, we see that if When = tree, BBE terminates when $sol = nil$.
This happens only when the search fails to find a solution of the current 
problem with a preference strictly greater than the current lower bound.
Let us denote with $Q^i$ and $Q^{i+1}$ 
respectively the IFCSPs given in input to the $i$-th and $i+1$-th 
recursive call of BBE.
First we notice that only procedure 
$Elicit@tree$ modifies the IFCSP in input 
by possibly adding new elicited preferences.
Moreover, whatever the value of parameter What is, 
the returned IFCSP is either the same as the one in input 
or it is a (possibly partial) completion of the one in input.
Thus we have $Q^{i+1} \in PC(Q^i)$ and $Q^{i} \in PC(P)$.
Since the search is always performed on the $\1$-completion of the current 
IFCSP, we can conclude that for every solution $s$, 
$pref(Q^{i+1},s) \leq pref(Q^{i},s)$. 
Let us now denote with $lb^i$ and $lb^{i+1}$ the lower bounds 
given in input respectively to the $i$-th and $i+1$-th 
recursive call of BBE. It is easy to see that $lb^{i+1} \geq lb^i$.
Thus, since at every iteration we have that the preferences of solutions 
can only get lower, and the bound can only get higher, and since 
we have a finite number of 
solutions, we can conclude that BBE always terminates. 

%Soundness and correctness
The reasoning that follows relies on the fact that value $pref$ returned 
by function $Elicit@tree$ is the final preference after elicitation 
of assignment $sol$ given in input. This is true since 
either What = all and thus all preferences have been elicited 
and the overall preference of $sol$ can be computed 
or only the worst preference has been elicited 
but in a fuzzy context where the overall 
preference coincide with the worst one.
If called with When = tree
IFCSP-SCHEME exits when the last branch and bound 
search has ended returning 
$sol = nil$.
In such a case $sol$ and $pref$ are updated to contain the 
best solution and associated preference found so far, i.e., $sol'$ and $pref'$.
Then, the algorithm returns the current IFCSP, say $Q$, and $sol$ and 
$pref$.
Following the same reasoning as above done for $Q^{i}$ we can conclude that 
$Q \in PC(P)$. 

At the end of every while loop execution, 
assignment $sol$ either contains  
an optimal solution $sol$ of the $\1$-completion of the current IFCSP 
or $sol = nil$. 
$sol=nil$ iff  there is no assignment with 
preference higher than $lb$ 
in the $\1$-completion of the current IFCSP. 
In this situation, $sol'$ and $pref'$ 
are an optimal solution and preference of the $1$-completion
of the current IFCSP.
However, since the preference of $sol'$, $pref'$ is 
independent of unknown preferences
and since due to monotonicity the optimal 
preference value of the 
$\1$-completion is always greater than or equal to 
that of the $\0$-completion we have that 
$sol'$ and $pref'$  are an optimal solution 
and preference of the $\0$-completion
of the current IFCSP as well.

By Theorems 1 and 2 of \cite{cp07-isoft} %\ref{theonos} and \ref{theonos0} of  ,???????????????????????????????
we can conclude that $NOS(Q)$ is not empty. 
If $pref= \0$, then 
$NOS(Q)$ contains all the assignments and thus also $sol$. 
The algorithm correctly returns 
the same IFCSP given in input, assignment $sol$ and its preference 
$pref$. If instead $\0 < pref$,
again the algorithm is correct, since by
Theorem 1 of \cite{cp07-isoft} %\ref{theonos} ???????
we know that $NOS(Q) = Opt(Q_0)$, and 
we have shown that $sol \in Opt(Q_0)$.  \\
%}

%\begin{theorem}
%Given a Fuzzy ISCSP $P$ and $when = branch$ or $when = node$
%in input, Algorithm {\em Find-NOS} always terminates, and it returns 
%an ISCSP $Q$ such that  $Q \in PC(P)$, an assignment $s \in NOS(Q)$, 
%and its preference in  $Q$.
%\end{theorem} 

%\proof{
%Termination
{\em Case 2:} When=branch or node.\\
In order to prove that the algorithm terminates, 
it is sufficient to show that $BBE$ terminates. 
Since the domains are finite, the labeling phase 
produces a number of finite choices at every level of the search tree. 
Moreover, since the number of variables is limited, 
then, we have also a finite number of levels in the tree.  
Hence, $BBE$ considers at most all the possible assignments, 
that are  a finite number. 
At the end of the execution of IFCSP-SCHEME, %{\em Find-NOS}, 
$sol$, with preference $pref$ is %the 
one of the optimal solutions of the current $P_1$%$P[?/1]$. 
%If $s_{max}\in Opt(P[?/1])$, it means that 
Thus,
for every assignment  $s'$, %$pref(P[?/1],s') \leq pref(P[?/1], sol)$. 
$pref(P_1,s') \leq pref(P_1, sol)$. 
Moreover, 
for every completion $Q'\in C(P)$ and for every assignment 
$s'$,   $pref(Q',s') \leq$ %pref(P[?/1], s')$. 
$pref(P_1, s')$. Hence, for every assignment $s'$ and for every 
$Q'\in C(P)$, we have that  $pref(Q',s') \leq$ %pref(P[?/1], sol)$. 
$pref(P_1, sol)$. In order to prove that $sol \in NOS(P)$, 
now it is sufficient to prove that 
for every $Q'\in C(P)$, %$pref(P[?/1], sol) = pref(Q', sol)$. 
$pref(P_1, sol) = pref(Q', sol)$. 
This is true, since $sol$ %\in Fixed(P)$ 
has a preference that is independent from the missing preferences of $P$, 
both when eliciting 
all the missing preferences, and when eliciting only the worst one 
either at branch or node level. 
In fact,  in both cases, the preference 
of $sol$ is the same in every completion. Q.E.D.\\
}
%-----------------

%While Unlike the algorithm in \cite{cp07-isoft} and, in particular, whenever 
If When=tree, then we elicit after each BB run, % of branch and bound,
and it is proven in \cite{cp07-isoft} 
that IFCSP-SCHEME never elicits preferences 
involved in solutions which are not possibly optimal.
This is a desirable property, 
since only possibly optimal solutions can become 
necessarily optimal.
However, the experiments will show % (see next section) 
that solvers satisfying such a desirable property are 
often out-performed in practice.

\section{Problem generator and experimental design}

To test the performance of these different
algorithms, we created IFCSPs using a generator which
is a simple extension of the standard random model for
hard constraints to soft and incomplete constraints. 
The generator has the following parameters:
\begin{itemize}
\item $n$: number of variables;
\item $m$: cardinality of the variable domains; %mirco la chiama $e$
\item $d$: density, that is,
the percentage of binary constraints
present in the problem w.r.t. the total number of possible binary
constraints that can be defined on $n$ variables;
\item $t$: tightness, that is,
the percentage of tuples with preference $0$ in each constraint 
and in each domain 
w.r.t. the total number of tuples ($m^2$ for the constraints, 
since we have only binary constraints, and $m$ in the domains);
\item $i$: incompleteness,
that is, the percentage of incomplete tuples (that is,
tuples with preference $?$) in each constraint and in each domain.
\end{itemize}
Given values for these parameters, we generate IFCSPs as follows.
We first generate $n$ variables and then $d$\% of the $n(n-1)/2$ possible constraints.
Then, for every domain and for every constraint,
we generate a random preference value in $(0,1]$ for each 
of the tuples (that are $m$ for the domains, and  $m^2$ for the constraints);
we randomly set $t$\% of these preferences to $0$;
and we randomly set $i$\% of the preferences as incomplete.

%For example, if the generator is given in input
%$n=10$, $m=5$, $d=50$, $t=10$, and $i=30$,
%it generates a binary IFCSP with $10$ variables,
%each with $5$ elements in the domain,
%$22$ constraints (on a total of $45 =n(n-1)/2$),
%$2$ tuples with preference $0$ and $7$ incomplete tuples (over a total of
%$25$) in each constraint, and $1$ missing preference in each domain.

%\subsection{Experimental design}

Our experiments %consider IFCSPs generated as above.  They 
measure the {\em percentage of elicited preferences} 
(over all the missing preferences) as the generation parameters vary.
Since some of the algorithm instances require the user 
to suggest the value for the next variable, 
we also show the {\em user's effort} in the various solvers, 
formally defined as the number %percentage 
of missing preferences the user has to consider 
to give the required help.

Besides the 16 instances of the scheme described above, we also 
considered a "baseline" algorithm that 
elicits preferences of randomly chosen tuples
every time branch and bound ends. All 
algorithms are named by means of the three parameters.
For example, algorithm DPI.WORST.BRANCH has parameters 
Who=dpi, What=worst, and When=branch.
For the baseline algorithm, we use the name DPI.RANDOM.TREE. 

For every choice of parameter values, 100 problem instances 
are generated. The results shown are the average 
over the 100 instances. 
Also, when it is not specified otherwise, we set
$n=10$ and $m=5$.
%Using 10 variables is sufficient to provide significant data.
However, we have similar results (although not shown in this paper for lack of space) 
for $n$ = 5, 8, 11, 14, 17, and 20.
All our experiments have been performed on an 
AMD Athlon 64x2 2800+, with 1 Gb RAM, Linux operating system, and using JVM 6.0.1.

\section{Results}

In this section we summarize and discuss our experimental 
comparison of the different algorithms. 
We first focus on incomplete fuzzy CSPs. We then consider 
two special cases: incomplete CSPs where all constraints
are hard, and incomplete fuzzy temporal problems.
In all the experimental results, the association between 
an algorithm name and a line symbol is shown below. %: %
%in Figure \ref{names}.

%\begin{figure}[h]
{\centering
{
\includegraphics[width=1\linewidth]{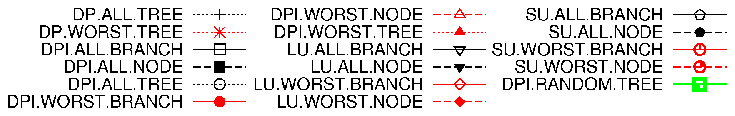}}
}
\vspace*{-.5cm}
%\caption{Names of the algorithms and corresponding line symbols.}
%\label{names}
%\end{figure}

%{\centering
%{\includegraphics[width=1\linewidth]{./grafici/grafici/aaai08_legenda.eps}}
%}
%\vspace*{-.3cm}

\subsection{Incomplete fuzzy CSPs}

%The names of all the %considered 
%algorithms and the corresponding line symbols are shown in Figure \ref{names}. 

\begin{figure*}[htbp]
\centering
\subfigure[d=50\%, t=10\%]
{\includegraphics[width=0.45\linewidth]{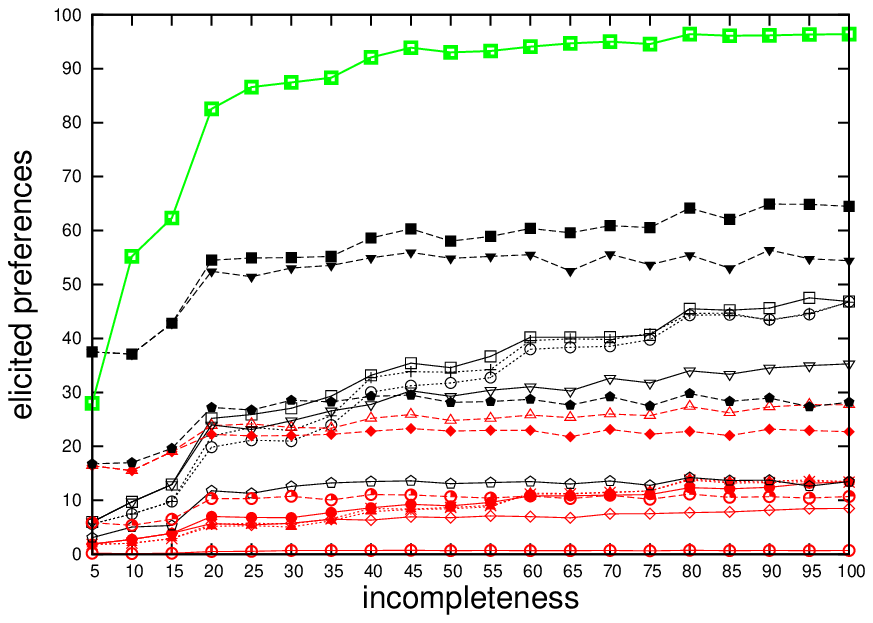}}
 \hspace{0mm}
 \subfigure[t=35\%, i=30\%]    {\includegraphics[width=0.45\linewidth]{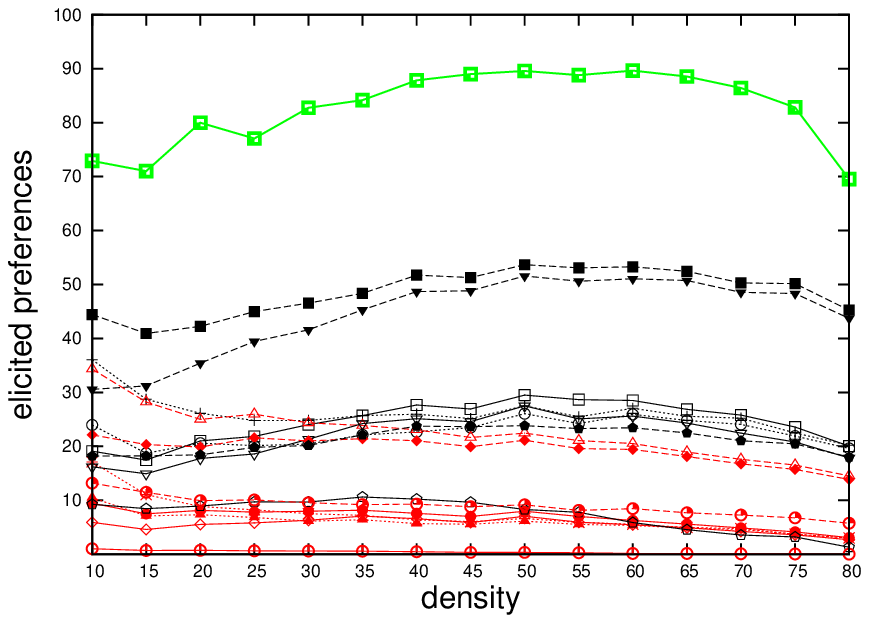}}
 \hspace{0mm}
 \subfigure[d=50\%, i=30\%]
{\includegraphics[width=0.45\linewidth]{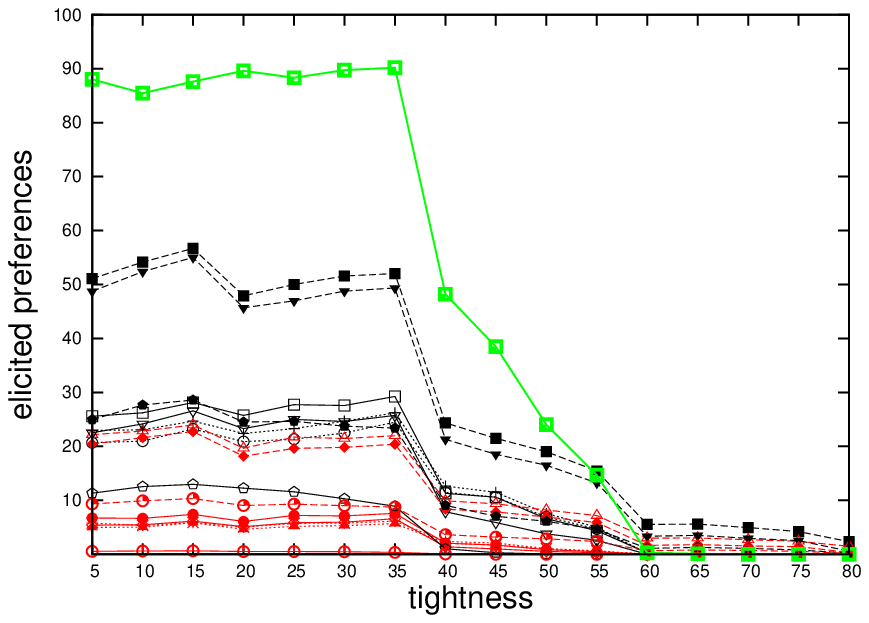}}
\vspace*{-.5cm}
\caption{Percentage of elicited preferences in incomplete fuzzy CSPs.} % with n=10 and e=5.}
\label{ris1}
\end{figure*}

Figure \ref{ris1} shows the percentage of elicited preferences 
when we vary the incompleteness, 
the density, and the tightness  respectively. 
For reasons of space, we show only 
the results for specific values of the parameters.
However, the trends observed here hold 
in general. 
It is easy to see that the best algorithms are those that 
elicit at the branch level. In particular, algorithm SU.WORST.BRANCH
elicits a very small percentage of missing preferences (less than 5\%), 
no matter the amount of incompleteness in the problem,
and also independently of the density and the tightness. 
This algorithm outperforms all others, but relies on help from the user.
The best algorithm that does not need such help is DPI.WORST.BRANCH.
This never elicits more than about 10\% of the missing preferences.
Notice that the baseline algorithm 
is always the worst one, and needs nearly all the missing preferences 
before it finds a necessarily optimal solution.
Notice also that the algorithms with What=worst 
are almost always better than those with What=all, and that
When=branch is almost always better than When=node or When=tree.

\begin{figure*}[htbp]
 \centering
 \subfigure[d=50\%, t=10\%]
   {\includegraphics[width=0.45\linewidth]{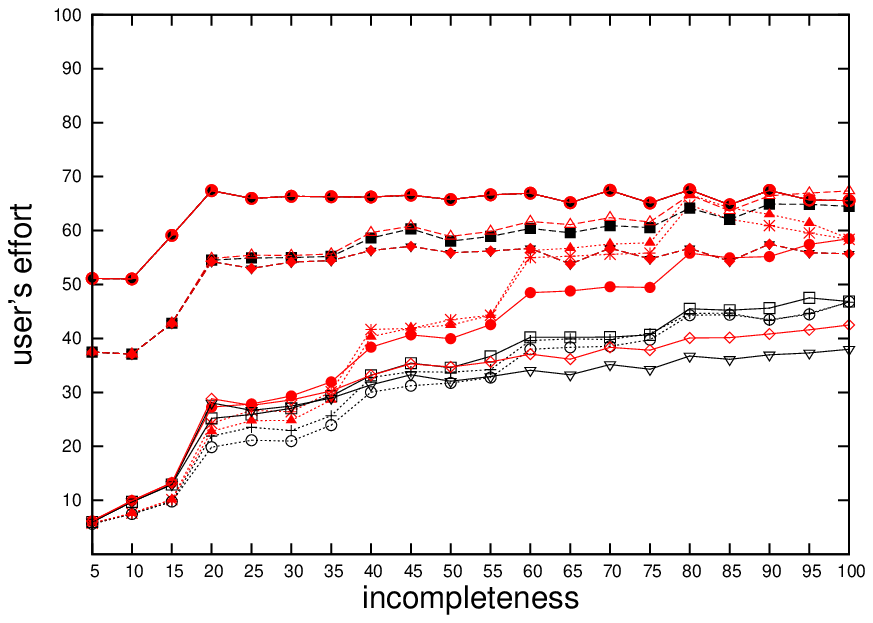}
}
\hspace{0mm}
  \subfigure[t=10\%, i=30\%]
{\includegraphics[width=0.45\linewidth]{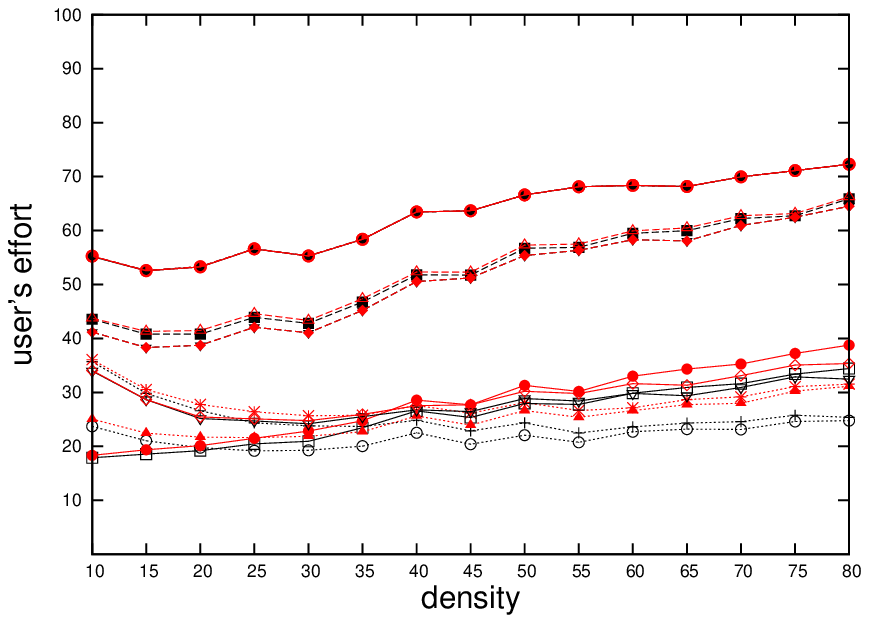}}
% \subfigure[t=10\%, i=30\%]
%   {\includegraphics[width=0.3\linewidth]{./grafici/grafici/aaai08_dpi_worst_branch.eps}}
% \hspace{0mm}
% \subfigure[d=50\%, t=10\%]
%   {\includegraphics[width=0.3\linewidth]{./grafici/grafici/aaai08_lu_worst_branch.eps}}
\vspace*{-.5cm}
\caption{Incomplete fuzzy CSPs: user's effort}  %and solution quality.} %best algorithms.}
\label{ris2}
\end{figure*}

Figure \ref{ris2} (a) shows the user's effort
as incompleteness varies. 
As could be predicted, the 
effort grows slightly with the incompleteness level, 
and it is equal to the percentage of elicited preferences 
only when What=all and Who=dp or dpi.
For example, when What=worst, even if Who=dp or dpi,
the user has to consider more preferences than those elicited, since 
to identify the worst preference value the user needs to check all of them
(that is, those involved in a partial or complete assignment).
DPI.WORST.BRANCH 
requires the user to look at 
60\% of the missing preferences at most, even when 
incompleteness is 100\%.

Figure \ref{ris2} (b) shows the user's effort
as density varies. Also in this case, as expected, the 
effort grows slightly with the density level.  
In this case DPI.WORST.BRANCH 
requires the user to look at most 
40\% of the missing preferences, even when 
the density is 80\%.

\begin{figure*}[htbp]
 \centering
% \subfigure[d=50\%, t=10\%]
%   {\includegraphics[width=0.45\linewidth]{./grafici/grafici/aaai08_100_F_usereffF_inc_5-100totale.eps}
%\hspace{0mm}
 \subfigure[solution quality]
{\includegraphics[width=0.45\linewidth]{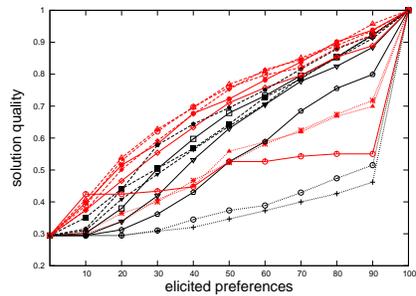}}
% \subfigure[d=50\%, t=10\%]
%   {\includegraphics[width=0.3\linewidth]{./grafici/grafici/aaai08_dpi_worst_branch.eps}}
% \hspace{0mm}
% \subfigure[d=50\%, t=10\%]
%   {\includegraphics[width=0.3\linewidth]{./grafici/grafici/aaai08_lu_worst_branch.eps}}
\vspace*{-.5cm}
\caption{Incomplete fuzzy CSPs:  solution quality.} %best algorithms.}
\label{ris3}
\end{figure*}

All these algorithms have a useful anytime property, since 
they can be stopped even before their termination
obtaining a possibly optimal solution 
with preference value equal to the
best solution considered up to that point.
Figure \ref{ris3} shows how fast the various algorithms reach optimality.
The $y$ axis represents the solution quality during execution, normalized 
to allow for comparison among different problems.
The algorithms that perform best in terms 
of elicited preferences, such as DPI.WORST.BRANCH, 
are also those that approach optimality fastest. 
We can therefore stop 
such algorithms early and still obtain a solution 
of good quality in all completions.

\begin{figure*}[htbp]
 \centering
 \subfigure[d=50\%, t=10\%]
   {\includegraphics[width=0.45\linewidth]{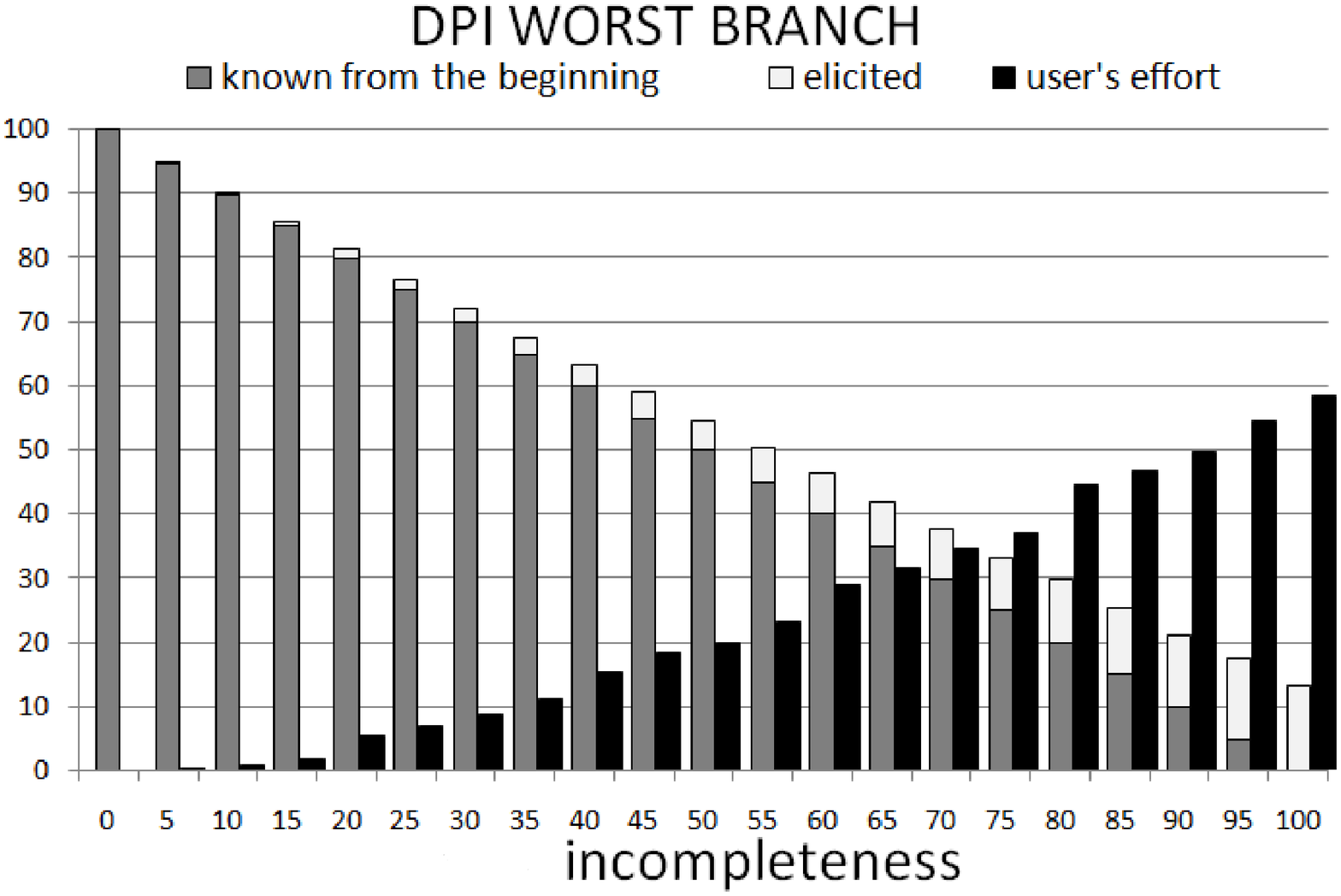}}
 \hspace{0mm}
 \subfigure[d=50\%, t=10\%]
   {\includegraphics[width=0.45\linewidth]{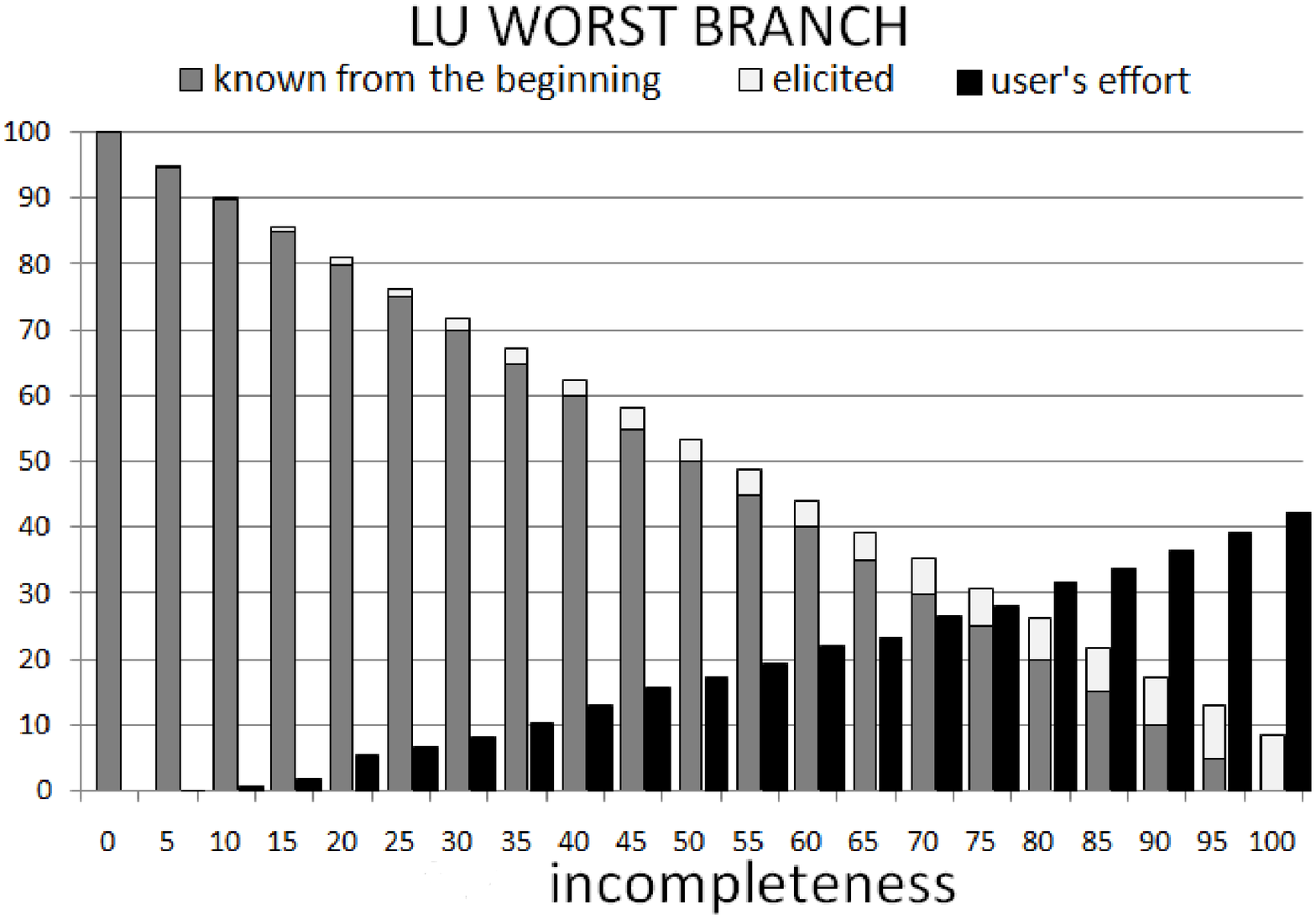}}
  \vspace*{-.5cm} 
\caption{Incomplete fuzzy CSPs: best algorithms.}
\label{best-fuzzy}
\end{figure*}

Figure \ref{best-fuzzy} (a) shows 
the percentage of elicited preferences over all the preferences (white bars) and the 
user's effort (black bars), as well as the percentage of preferences 
present at the beginning (grey bars) for DPI.WORST.BRANCH.
Even with high levels of incompleteness,
this algorithm elicits only a very small fraction of the preferences, 
while asking the user to consider at most 
half of the missing preferences.

Figure \ref{best-fuzzy} (b) shows results for 
LU.WORST.BRANCH, where the user is involved in the choice of the value 
for the next variable. Compared to DPI.WORST.BRANCH, this algorithm 
is better both in terms of elicited preferences and user's effort
(while SU.WORST.BRANCH is better only for the elicited preferences).
We conjecture that the help the user gives in choosing the next value 
guides the search towards better solutions, thus resulting in an overall decrease
of the number of elicited preferences. 

Although we are mainly interested in the amount of elicitation,
we also computed the time to run the 16 algorithms.
Ignoring the time taken to ask the user for missing preferences,
the best algorithms need about 200 ms to find
the necessarily optimal solution
for problems with 10 variables and 5 elements in the domains,
no matter the amount of incompleteness. 
Most of the algorithms need less than 500 ms.

%\begin{figure}[htbp]
%\centering
%{\includegraphics[width=0.5\linewidth]{./grafici/grafici/solquality100.eps}}
%\vspace*{-.5cm}
%\caption{Solution quality vs. \% of elicited preferences.}
%\label{qual}
%\end{figure}

%-------

\subsection{Incomplete hard CSPs}

%hard

\begin{figure*}[htbp]
 \centering
 \subfigure[d=50\%, t=10\%]
   {\includegraphics[width=0.45\linewidth]{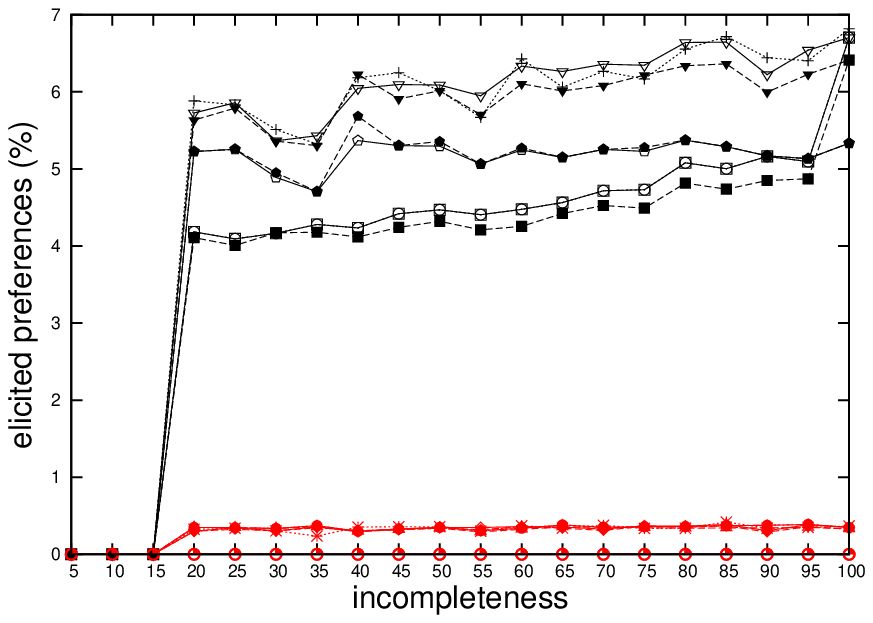}
}
 \hspace{0mm}
 \subfigure[t=10\%, i=30\%]
   {\includegraphics[width=0.45\linewidth]{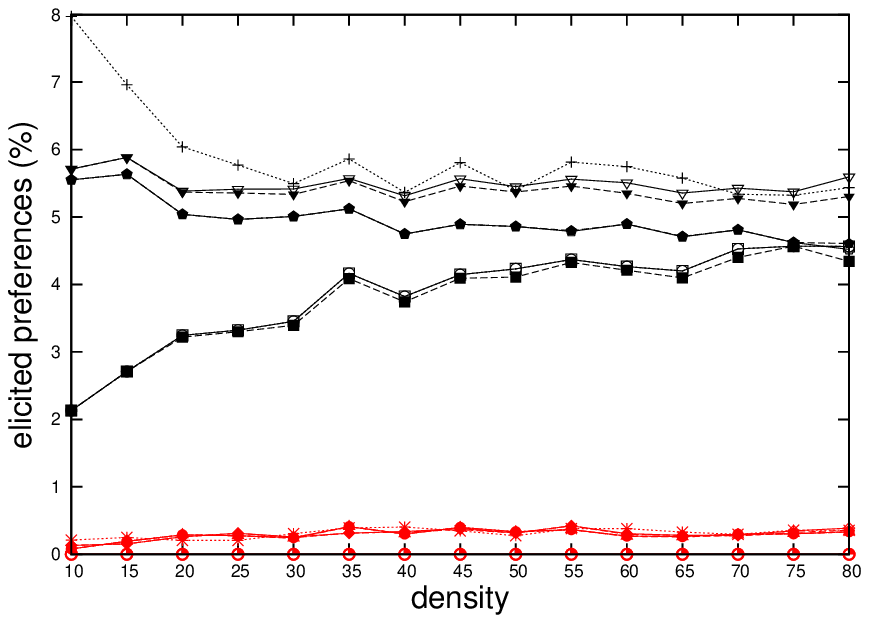}
}
 \hspace{0mm}\\
 \subfigure[d=50\%, i=30\%]
   {\includegraphics[width=0.45\linewidth]{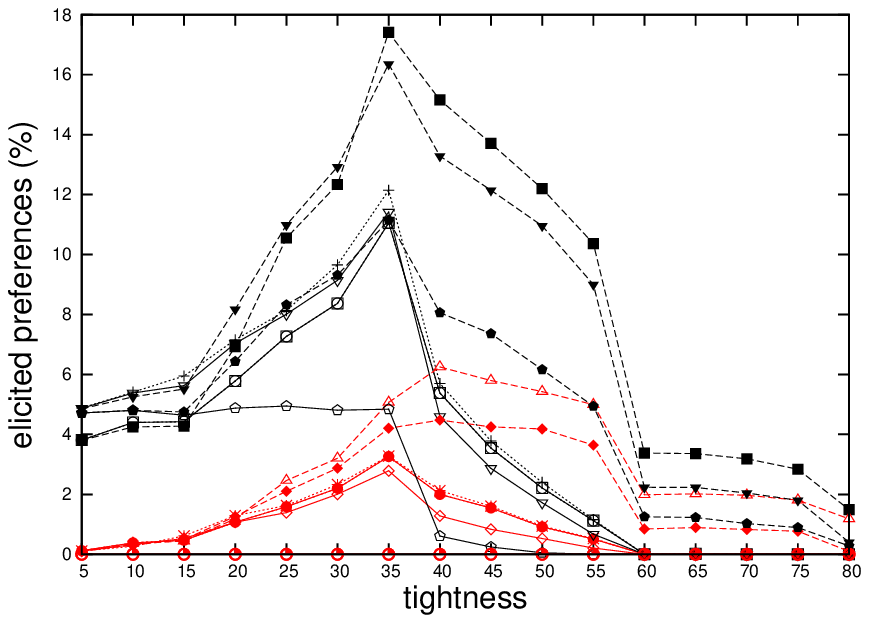}}
 \vspace*{-.5cm}  
\caption{Elicited preferences in incomplete CSPs.} % with n=10 and e=5.}
\label{ris4}
\end{figure*}

\begin{figure*}[htbp]
 \centering
 \subfigure[d=50\%, t=10\%]
   {\includegraphics[width=0.45\linewidth]{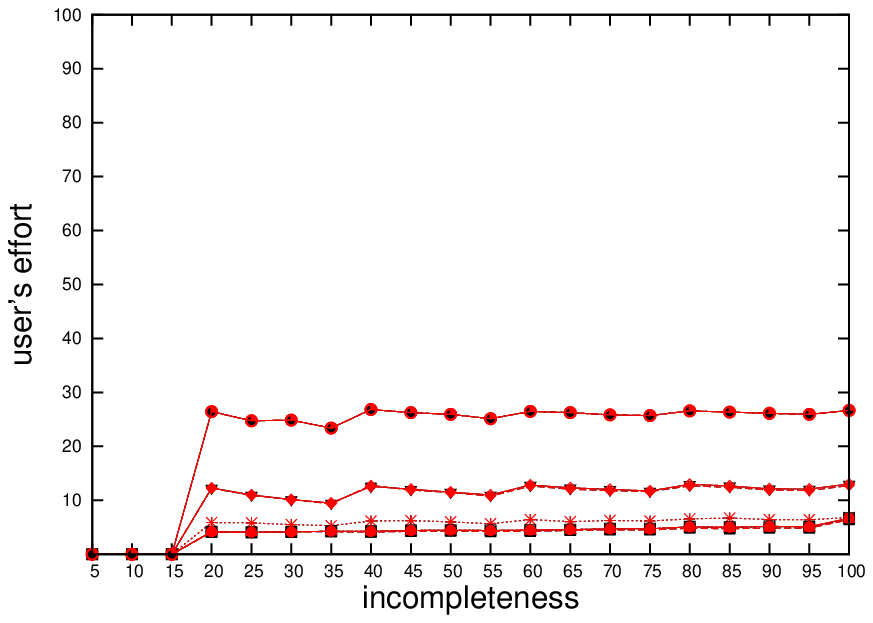}
}
 \hspace{0mm}
 \subfigure[t=10\%, i=30\%]
 {\includegraphics[width=0.45\linewidth]{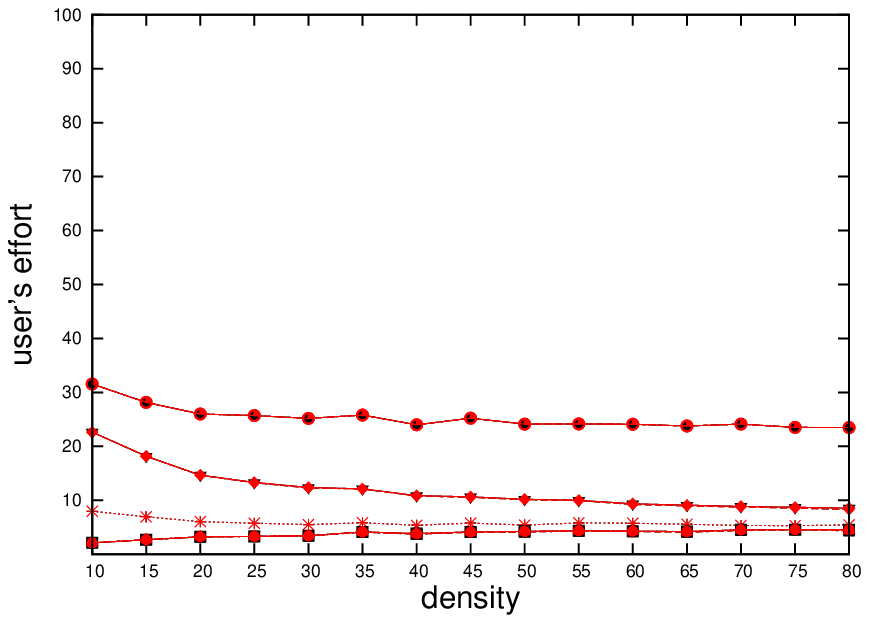}}
 \vspace*{-.5cm}
\caption{Incomplete CSPs: user's effort} %and (b) best algorithm.}
\label{ris5}
\end{figure*}

\begin{figure*}[htbp]
 \centering
% \subfigure[d=50\%, t=10\%]
%   {\includegraphics[width=0.45\linewidth]{./grafici/grafici/aaai08_100_H_usereffH_inc_5-100totale.eps}
%}
% \hspace{0mm}
 \subfigure[d=50\%, t=10\%]
 {\includegraphics[width=0.45\linewidth]{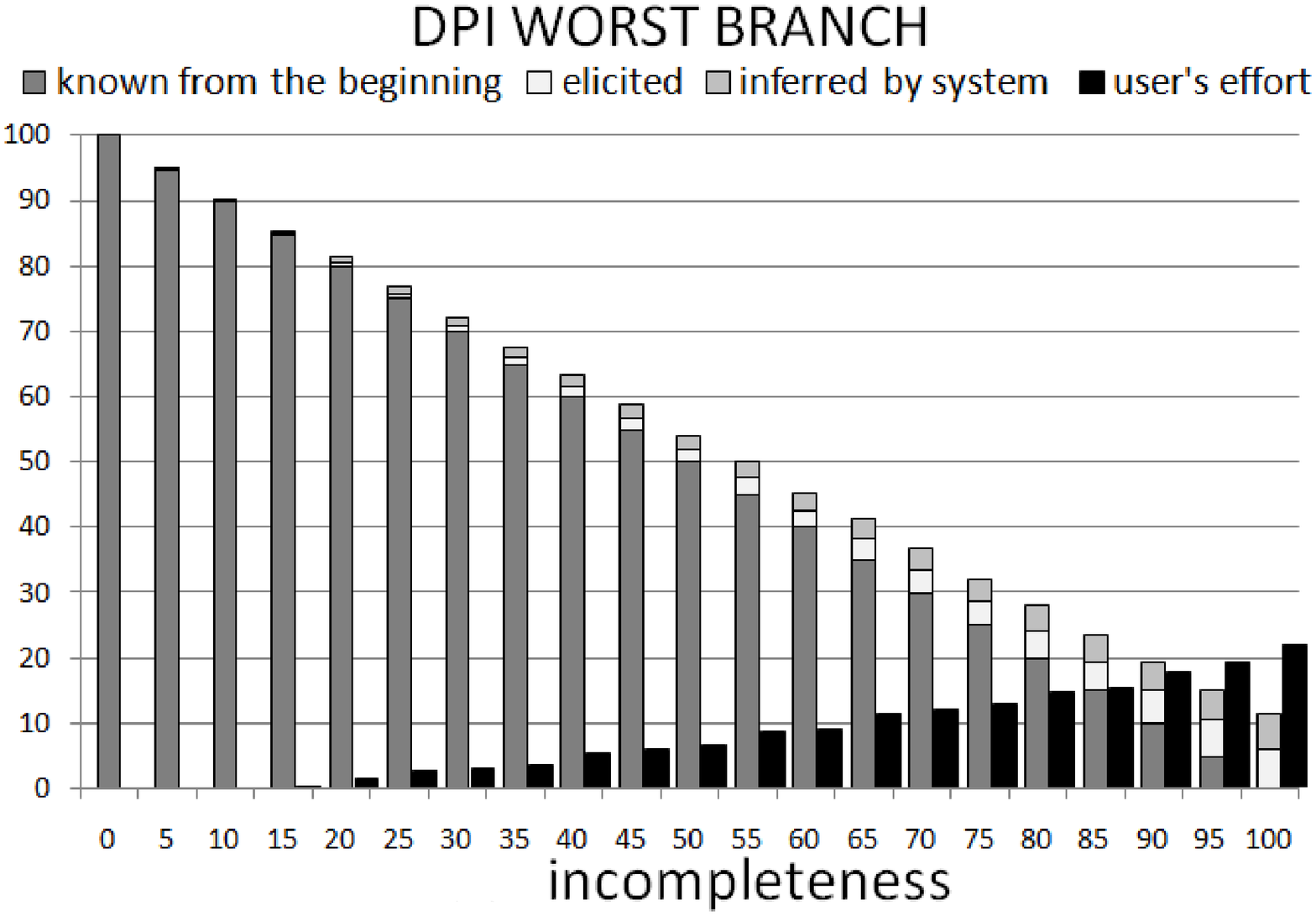}}
\vspace*{-.5cm}
\caption{Incomplete CSPs: best algorithm.}
\label{ris6}
\end{figure*}

We also tested these algorithms on hard CSPs. In this case, 
preferences are only 0 and 1, and necessarily optimal solutions
are complete assignments which are feasible in all completions. 
The problem generator is adapted accordingly.
The parameter What now has a specific meaning: What=worst
means asking if there is a 0 in the missing preferences. 
If there is no 0, we can infer that all the missing preferences are 1s. 
%Such a kind of inference is very useful, as shown has shown to be 

Figure \ref{ris4} shows the percentage of elicited preferences for hard CSPs 
in terms of amount of incompleteness, density, and tightness.
Notice that the scale on the $y$ axis varies to include only the 
highest values. 
The best algorithms are those with What=worst, where the inference 
explained above about missing preferences can be performed. 
It is easy to see a phase transition 
at about 35\% tightness, 
which is when problems pass from being solvable 
to having no solutions.
However, the percentage of elicited preferences is below 20\% for all 
algorithms even at the peak.

Figure \ref{ris5} (a) shows the user's effort in terms of amount of incompleteness and Figure \ref{ris5} (b) shows  the user's effort in terms of density for the case of hard CSPs.
Overall, the best algorithm is again DPI.WORST.BRANCH.
Figure \ref{ris6} gives the elicited preferences and user 
effort for this algorithm.

\subsection{Incomplete temporal fuzzy CSPs}

\begin{figure}
\centering
% \subfigure[Temporal SCSP, n=10, de=30, r=7, a=5, b=6, c=7, i=30\%]
%   {\includegraphics[width=0.3\linewidth]{./grafici/aaai08_100_T_F_percincT_F_den_10-80totale.eps}}
% \hspace{0mm}
%\subfigure[Temporal SCSP, n=10, d=50\%, de=30, r=7, a=5, b=6, c=7]
{\includegraphics[width=0.5\linewidth]{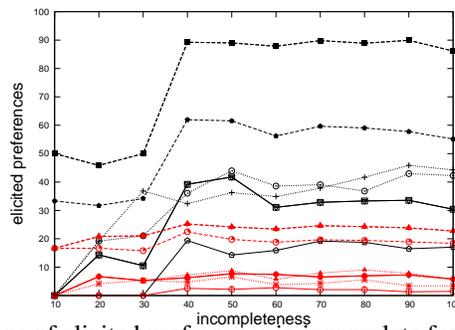}}
\vspace*{-.5cm}
\caption{Percentage of elicited preferences in incomplete fuzzy temporal CSPs.}
\label{temp}
\end{figure}

We also performed some experiments on fuzzy simple temporal problems 
\cite{stpp}.
These problems have constraints of the form $a \leq x-y \leq b$ modelling allowed 
time intervals for durations and distances of events, and fuzzy preferences  
associated to each element of an interval.
We have generated classes of such problems following the approach in \cite{stpp},
adapted to consider incompleteness.
While the class of problems generated in \cite{stpp} is tractable,  
the presence of incompleteness makes them intractable in general.
Figure \ref{temp} shows that in this specialized domain it is also possible to 
find a necessarily optimal solution by asking about 10\% of the missing preferences,
for example via algorithm DPI.WORST.BRANCH.

\section{Future work}

In the problems considered in this papers, we have no information about the missing 
preferences. We are currently considering settings in which each missing preference  
is associated to a range of possible values, that may be smaller than the whole range of 
preference values. For such problems, we 
intend to define several notions of optimality, among which 
necessarily and possibly
optimal solutions are just two examples, and to develop specific
elicitation strategies for each of them.
We are also studying soft constraint problems when no preference is missing, but some of them 
are unstable, and have associated a range of possible alternative values.

To model fuzzy CSPs, we have not used traditional fuzzy set theory \cite{dub80}, but soft CSPs \cite{jacm}, since we intend to apply our work also to non-fuzzy CSPs. 
%Some of our algorithms work with non-fuzzy CSPs (e.g., those with What=all). A general formulation therefore permits us to analyse such scenarios.  A degree of preference is a measure of satisfaction and not uncertainty. In this paper, the only form of uncertainty is an absence of some preference values. We intend to consider type-2 fuzzy logic systems \cite{type2} in future to model other types of uncertainty.
In fact,  we plan to consider incomplete weighted constraint problems as well as different heuristics for choosing the next variable during the search.
All algorithms with What=all are not tied to fuzzy CSPs and are reasonably efficient. 
%We are confident therefore to have good results with other classes of CSPs.
Moreover, we intend to build solvers based on local search and variable elimination methods.
Finally, we want to add elicitation costs and to use 
them also to guide the search, as done in \cite{nic} for hard CSPs.

\section*{Acknowledgements}

This work has been partially supported by Italian MIUR PRIN project
``Constraints and Preferences'' (n. $2005015491$). 
The last author is funded by the
Department of Broadband, Communications and the Digital Economy, 
and the Australian Research Council.

\bibliography{biblio-isoft2}
\end{document}